\documentclass[runningheads]{llncs}
\usepackage[T1]{fontenc}
\usepackage{subcaption} 
\usepackage{pgfplots}
\pgfplotsset{compat=1.18}
\usepackage{sansmath}
\usetikzlibrary{patterns}
\usepackage{caption} 

\usepackage{bbding}
\usepackage{amsmath}
\usepackage{amssymb} 
\usepackage{graphicx}
\usepackage{array}
\usepackage{booktabs}  
\usepackage{algorithm}  
\usepackage{algpseudocode}  
\usepackage{hyperref}
\usepackage{multirow}
\usepackage{color}

\renewcommand{\geq}{\geqslant}

\begin{document}
\title{URAG: Implementing a Unified Hybrid RAG for Precise Answers in University Admission Chatbots – A Case Study at HCMUT}

%

\titlerunning{URAG: A Unified RAG for Precise University Admission Chatbots}
%
\author{Long Nguyen \orcidID{0009-0008-7488-4714} \and 
Tho Quan\orcidID{0000-0003-0467-6254}}
\authorrunning{Long Nguyen et al.}
%
\institute{URA Research Group, Faculty of Computer Science and Engineering, Ho Chi Minh City University of Technology (HCMUT), Ho Chi Minh City, Vietnam \and Vietnam National University Ho Chi Minh City,  Ho Chi Minh City, Vietnam \email{\{long.nguyencse2023,qttho\}@hcmut.edu.vn}}
\maketitle              
\begin{abstract}
With the rapid advancement of Artificial Intelligence, particularly in Natural Language Processing, Large Language Models (LLMs) have become pivotal in educational question-answering systems, especially university admission chatbots. Concepts such as Retrieval-Augmented Generation (RAG) and other advanced techniques have been developed to enhance these systems by integrating specific university data, enabling LLMs to provide informed responses on admissions and academic counseling. However, these enhanced RAG techniques often involve high operational costs and require the training of complex, specialized modules, which poses challenges for practical deployment. Additionally, in the educational context, it is crucial to provide accurate answers to prevent misinformation, a task that LLM-based systems find challenging without appropriate strategies and methods.
In this paper, we introduce the Unified RAG (URAG) Framework, a hybrid approach that significantly improves the accuracy of responses, particularly for critical queries. Experimental results demonstrate that URAG enhances our in-house, lightweight model to perform comparably to state-of-the-art commercial models. Moreover, to validate its practical applicability, we conducted a case study at our educational institution, which received positive feedback and acclaim. This study not only proves the effectiveness of URAG but also highlights its feasibility for real-world implementation in educational settings.

\keywords{Question-Answering Systems  \and Retrieval-Augmented Generation \and University Admission
Chatbots.}
\end{abstract}
\section{Introduction}

\textit{Artificial Intelligence} (AI) has become a fundamental component of modern technological advancements, transforming industries across the board, including education \cite{MEMARIAN2023100022}. Among AI’s various applications, \textit{Natural Language Processing} (NLP) has proven particularly valuable in the development of chatbots aimed at assisting with university admissions and providing comprehensive institutional information \cite{Lende2016QuestionAS}. These chatbots play an essential role in enhancing communication between universities and prospective students, ensuring that inquiries are met with timely and accurate responses.

Recent breakthroughs in AI, such as the introduction of the Attention Mechanism \cite{Vaswani2017AttentionIA} and the rise of \textit{Large Language Models} (LLMs), have significantly improved the performance of these educational chatbots \cite{aichatbot}. LLMs, with their ability to generate human-like text and perform complex tasks \cite{Naveed2023ACO}, offer users a more interactive and dynamic experience. However, despite these advancements, a critical challenge remains as LLM-based chatbots are prone to generating inaccurate or misleading responses, especially when handling specialized or context-specific queries. This issue, commonly referred to as \textit{hallucination} \cite{10.1145/3571730}, poses significant risks, particularly in high-stakes contexts like university admissions, where the accuracy of information regarding application deadlines or program details is paramount.

To mitigate these risks, the \textit{Retrieval-Augmented Generation} (RAG) approach has emerged as a potential solution \cite{10.5555/3495724.3496517}. RAG combines retrieval-based mechanisms with generative models, enabling chatbots to consult external sources of information before generating responses. While this approach helps reduce hallucinations and improves accuracy, early implementations of RAG have faced limitations, such as noise in retrieval results, a disconnect between retrieval and generation processes, and difficulties in managing longer contexts \cite{Zhao2024RetrievalAugmentedGF}. These limitations can still result in inaccurate responses and hallucinations. Furthermore, more advanced RAG systems \cite{Asai2023SelfRAGLT,Jeong2024AdaptiveRAGLT,Wang2024SpeculativeRE,Yan2024CorrectiveRA}, though promising, often introduce greater complexity and operational costs, making their deployment in real-world educational settings less feasible.

In response to these challenges, we propose the \textit{\textbf{U}nified \textbf{RAG}} (URAG) framework, specifically designed to improve lightweight LLMs for use in university admission chatbots. URAG integrates the reliability of rule-based systems with the adaptability of RAG, creating a two-tiered approach. The first tier leverages a comprehensive \textit{Frequently Asked Questions} (FAQ) system to provide accurate responses to common queries, especially those involving sensitive or critical information. If no match is found in the FAQ, the second tier retrieves relevant documents from an augmented database and generates a response through an LLM. 

\begin{figure}[!ht] \centering \includegraphics[width=\textwidth]{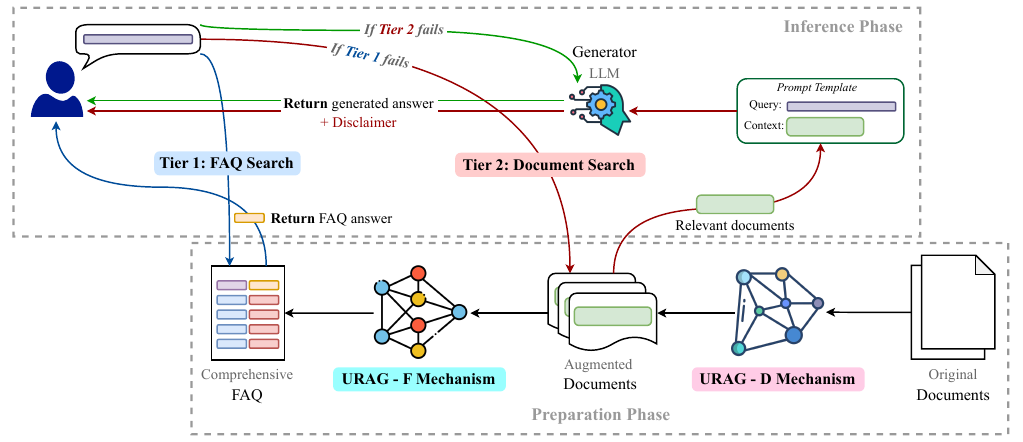} 
 \caption{The architecture of URAG framework illustrating the two-tiered approach for improving LLM performance in university admission chatbots.} 
 \label{URAG_Architecture} 
 \end{figure}

To enhance this process, we propose two key mechanisms, URAG-D for augmenting the original document database and URAG-F for generating an enhanced FAQ. These mechanisms not only enrich the database but also improve the retrieval process in both tiers, as shown in Figure \ref{URAG_Architecture}.

Our experiments highlight the effectiveness of URAG when paired with an in-house developed Vietnamese lightweight LLM. We benchmarked URAG’s performance against \textit{state-of-the-art} (SOTA) commercial chatbots, including GPT-4o\footnote{\url{https://openai.com/chatgpt/}}, Gemini 1.5 Pro\footnote{\url{https://gemini.google.com/}}, and Claude 3.5 Sonnet\footnote{\url{https://claude.ai/}} - models renowned for their vast parameter scales, encompassing trillions of parameters and training on extensive real-time data \cite{RAY2023121}. To further validate URAG’s practical application, we integrated it into HCMUT Chatbot\footnote{\url{https://www.ura.hcmut.edu.vn/bk-tvts/}}, where it has since gained recognition for its positive impact on university admissions at \textit{Ho Chi Minh City University of Technology} (HCMUT). 

In summary, our contributions are as follows.
\begin{itemize}
    \item We introduced URAG, a hybrid system that integrates rule-based and RAG approaches to enhance the performance of lightweight LLMs tailored for educational chatbots.
    \item We collected a real-world dataset of university admission questions from high school students and conducted a comprehensive evaluation of URAG against leading commercial chatbots, demonstrating its competitive performance.
    \item We successfully implemented URAG in a practical deployment at HCMUT, showcasing its effectiveness through a functional product that continues to address the university’s needs.
\end{itemize}

\section{Related Work}
\subsection{Retrieval-Augmented Generation (RAG)}
\begin{figure}[!ht] \centering \includegraphics[width=0.95\linewidth]{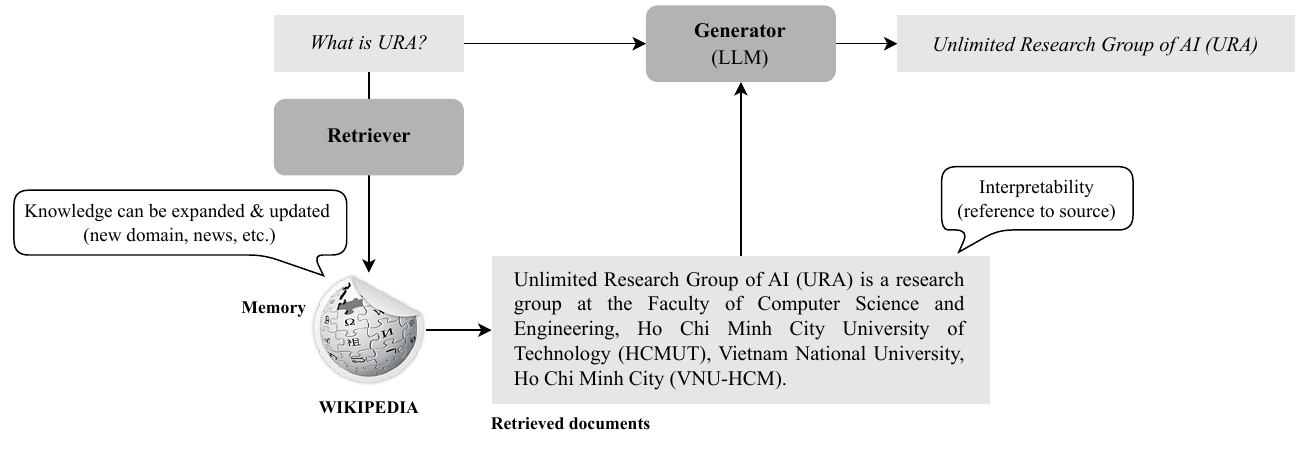} 
\caption{An example of a typical RAG pipeline.} \label{RAG} 
\end{figure}

RAG has become a widely adopted technique for building \textit{question-answering} (QA) systems using LLM. This approach is favored for its cost-efficiency and ability to combine retrieval-based methods with the generative power of LLMs. A typical RAG pipeline consists of two primary components, the Retriever and the Generator, as shown in Figure \ref{RAG}. One of the earliest forms of RAG, termed \textit{Naive RAG}, gained popularity following the release of models like ChatGPT \cite{Gao2023RetrievalAugmentedGF}.


Naive RAG operates by replacing the user's query with a prompt enriched by documents retrieved through the Retriever, which is then fed into the Generator (LLM) to generate the final output. This method, while straightforward, has proven to be an effective baseline for many LLM-powered QA systems.

\subsection{Strategies for Enhancing a RAG Pipeline}

Naive RAG implementations, while effective and straightforward, face significant limitations in complex QA systems, such as university admission chatbots \cite{Gao2023RetrievalAugmentedGF}. To address these challenges, more sophisticated RAG pipelines have emerged, with enhancements generally categorized into five key areas, such as input refinement, retriever improvements, generator optimization, result validation, and overall pipeline enhancements \cite{Zhao2024RetrievalAugmentedGF}.

Enhancing retrievers using structured data, like knowledge graphs, boosts precision and reliability by leveraging the inherent relationships within the data \cite{10.1145/3643479.3662055}. Techniques such as \textit{Graph RAG} (GRAG) \cite{Hu2024GRAGGR} employ graph topology, utilizing subgraph structures to improve information retrieval. However, maintaining such structured databases is particularly challenging in dynamic fields like education, where data is constantly evolving.

Further advancements in RAG pipelines include approaches like \textit{Self-Reflective RAG} (Self-RAG) \cite{Asai2023SelfRAGLT}, \textit{Corrective RAG} (CRAG) \cite{Yan2024CorrectiveRA}, and \textit{Adaptive RAG} \cite{Jeong2024AdaptiveRAGLT}. These methods enrich inputs by selecting and supplementing relevant information, often incorporating web searches before feeding the data into the LLM. Additionally, these techniques can detect and correct hallucinations or inaccuracies, allowing for response regeneration to enhance accuracy. Another notable approach, \textit{Speculative RAG} \cite{Wang2024SpeculativeRE}, generates multiple response drafts by combining the query with relevant document clusters, processing them in parallel, and selecting the best answer through evaluation.

Despite these improvements in accuracy, these advanced RAG techniques introduce additional complexity. Iterative feedback loops required for response refinement lead to longer processing times and increased computational demands. Although speculative execution aims to reduce latency, the need for separate module training makes these sophisticated RAG models resource-intensive and challenging to implement in practical educational settings.

\subsection{Practical Approaches to University Admission Chatbots}

University admission chatbots have become vital tools in educational settings, providing immediate support to prospective students. Most of these systems are built around two core components, namely \textit{Natural Language Understanding} (NLU) and \textit{Natural Language Generation} (NLG). NLU processes like intent classification and entity recognition enable chatbots to interpret user queries, while NLG generates responses through rule-based approaches, predefined scripts, or machine learning techniques \cite{Aloqayli2023IntelligentCF}.

Many chatbots, such as those discussed in \cite{9648677,NGUYEN2021100036}, use platforms like \textit{Rasa} \cite{Bocklisch2017RasaOS}, which offer robust tools for training NLU models and managing dialogues. While these systems effectively handle frequently asked questions, they often struggle with complex or out-of-scope queries that are beyond their training data. To address these limitations, advanced models incorporate Sequence-to-Sequence architectures with Attention Mechanisms \cite{CHANDRA2019367} or simplified RAG structures fine-tuned on domain-specific datasets \cite{llamachatbot,Neupane2024FromQT}, ensuring that responses are generated continuously to maintain user engagement, even if they lack complete accuracy.

Additionally, some chatbots integrate third-party APIs like GPT-3.5 Turbo \cite{Nguyen2022AIPoweredUD,10.1145/3627508.3638293}, enhancing their capabilities but introducing significant concerns. These integrations pose security risks due to external data handling \cite{zeng-etal-2024-good} and lead to high long-term costs, which can be unsustainable for many institutions.

Despite these advancements, relying heavily on LLMs still results in challenges, such as inaccuracies in responses to sensitive or context-specific queries. Therefore, there is a pressing need for a streamlined solution that balances the power of LLMs with efficient, lightweight designs and incorporates mechanisms to ensure accuracy for critical responses. Such a solution should provide reliable performance tailored to the specific requirements of university admission inquiries, addressing the gaps in current chatbot technologies.

\section{URAG: A Unified RAG for Precise University Admission Chatbots}

University admission chatbots frequently encounter challenges in providing accurate responses to common questions, particularly those involving critical details like admission requirements, deadlines, scores, and department codes. Such information requires high precision, as inaccuracies can mislead prospective students. Traditionally, human advisors rely on FAQ scripts to ensure consistency and accuracy. When questions fall outside the FAQ's coverage, advisors consult additional documents to provide the correct information. 

Inspired by this conventional advisory approach, we introduce the URAG framework, a two-tiered architecture designed to emulate this method and significantly improve chatbot performance.

\subsection{Overview of URAG Architecture}

\begin{figure}[!ht] \centering \includegraphics[width=\linewidth]{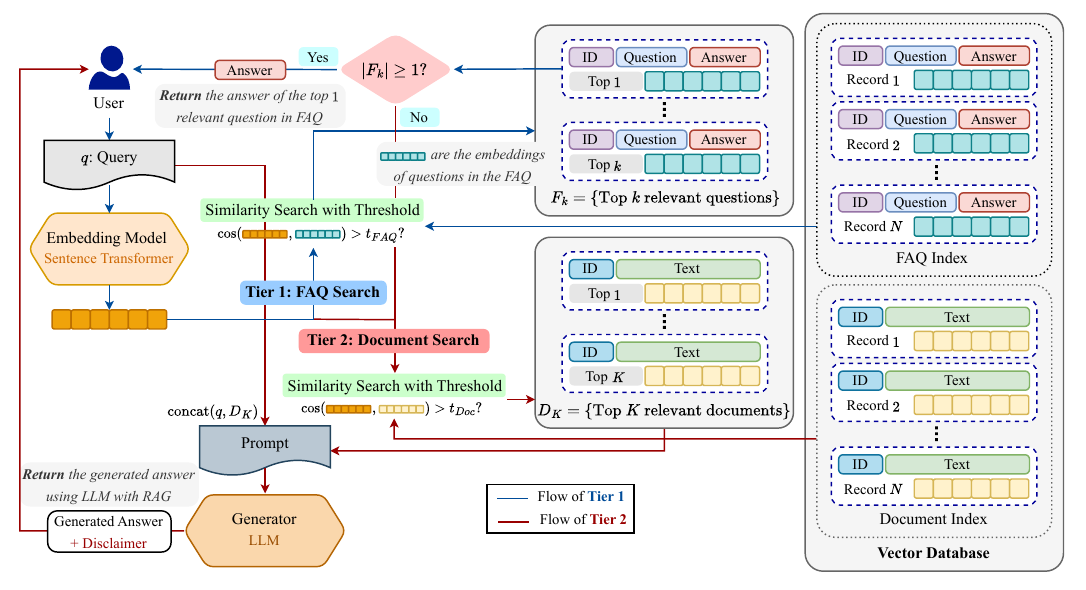} 
\caption{Illustration of the URAG framework, highlighting the two-tiered approach.} \label{URAG_General} 
\end{figure}

URAG operates on a unified dual-tier system, depicted in Figure \ref{URAG_General}.

\begin{description} 
\item[Tier 1] This tier utilizes an enriched FAQ database that extensively mines and integrates data from various text corpora with existing FAQs. The enrichment of this database is automated through the URAG-F mechanism, ensuring that the most common and critical inquiries receive direct and precise responses. 

\item[Tier 2] If no suitable match is found in the first tier, the process advances to the second tier. This tier searches within a document corpus that has been augmented by the URAG-D mechanism. It mimics the traditional RAG process by using a prompt template that guides the LLM to generate relevant responses based on the retrieved documents. 

\item[Fallback] If neither tier successfully retrieves the required information, the system defaults to generating a response directly from the LLM, accompanied by a disclaimer to manage user expectations about potential inaccuracies.
\end{description}


The operational flow of the URAG framework is systematically outlined in Algorithm \ref{alg:urag}, with key terms defined in Table \ref{tab:notation}.

\begin{table}[!hbt]
    \centering
    \caption{Notation}
    \label{tab:notation}
    \begin{tabular}{>{\raggedright\arraybackslash}p{2cm} p{10cm}}
        \toprule
        \textbf{Symbol} & \textbf{Description} \\
        \midrule
        $\mathbb{T}$ & The textual domain \\
        $\mathcal{E}:\mathbb{T}\to \mathbb{R}^m$ & A mapping function that transforms a text $t \in \mathbb{T}$ into an $m$-dimensional embedding vector $E(t) \in \mathbb{R}^m$, facilitated by an \textsf{Embedding Model} \\
       $\mathcal{L} : \mathbb{T} \to \mathbb{T}$ & A generation function that maps a prompt $p \in \mathbb{T}$ to a response $\mathcal{L}(p) \in \mathbb{T}$, generated by a \textsf{LLM} \\
        $f_i = \{a_i, b_i\}$ & A \textit{question-answer} (Q-A) pair from the enriched FAQ list $F$, where $a_i\in \mathbb{T}$ is a question and $b_i\in \mathbb{T}$ is its corresponding answer \\
        $d_i$ & A document segment within the augmented corpus $D$ \\
        $t_{\text{FAQ}}, t_{\text{Doc}}$ & \textbf{Thresholds} defining the acceptance criteria for FAQ and Document search tiers, respectively \\
        $w_i$ & A warning or disclaimer from a predefined set $W \subseteq \mathbb{T}$ \\
        \bottomrule
    \end{tabular}
\end{table}

\begin{algorithm}[!ht]
    \caption{URAG Operational Framework}
    \label{alg:urag}
    \begin{algorithmic}[1]
        \Statex \textbf{Input:} User query \( q \in \mathbb{T} \)
        \Statex \textbf{Output:} Final answer \( a_G \in \mathbb{T} \)
        \State \textbf{Query Embedding:} Compute \( q' = \mathcal{E}(q) \in \mathbb{R}^m \)
        \Statex \textbf{Tier 1: FAQ Search}
        \State Identify the top \( k \) FAQ pairs \( F_k = \{f_i\}_{i=1}^k \) such that each \( f_i = \{a_i, b_i\} \) satisfies \( \cos(\theta(q', \mathcal{E}(a_i))) \geq t_{\text{FAQ}} \), where \( a_i \in f_i \in F \) and \( i \in \{1, 2, .., k\} \)
        \If{\( |F_k| \geq 1 \)}
            \State Select answer \( b_j \) from the pair \( f_j = \{a_j, b_j\} \) with the \textbf{highest score} in $F_k$
            \State \textbf{Return} \( a_G = b_j \)
        \Else
            \State \textbf{Proceed to Tier 2}
        \EndIf
        \Statex \textbf{Tier 2: Document Search}
        \State Retrieve the top \( K \) document segments \( D_K = \{d_i\}_{i=1}^K \) that meet the threshold \( t_{\text{Doc}} \), meaning \( \cos(\theta(q', \mathcal{E}(d_i))) \geq t_{\text{Doc}} \), where \( d_i\in D\) and \( i \in \{1, 2, .., K\} \)
        \If{\( |D_K| \geq 1 \)}
            \State Construct prompt \( p = \text{concat}(q, D_K) \), where \( p \in \mathbb{T} \)
            \State Generate \( a_G = \mathcal{L}(p) \)
            \State \textbf{Return} \( a_G := \text{concat}(\mathcal{L}(p), w_s) \)
        \Else
            \State \textbf{Fallback:} Generate \( a_G = \mathcal{L}(q) \)
            \State \textbf{Return} \( a_G := \text{concat}(\mathcal{L}(q), w_r) \)
        \EndIf
    \end{algorithmic}
\end{algorithm}

\subsection{The Truth Behind ``Unified'' in URAG Architecture}

\begin{figure}[!ht]
    \centering
    \includegraphics[width=\linewidth]{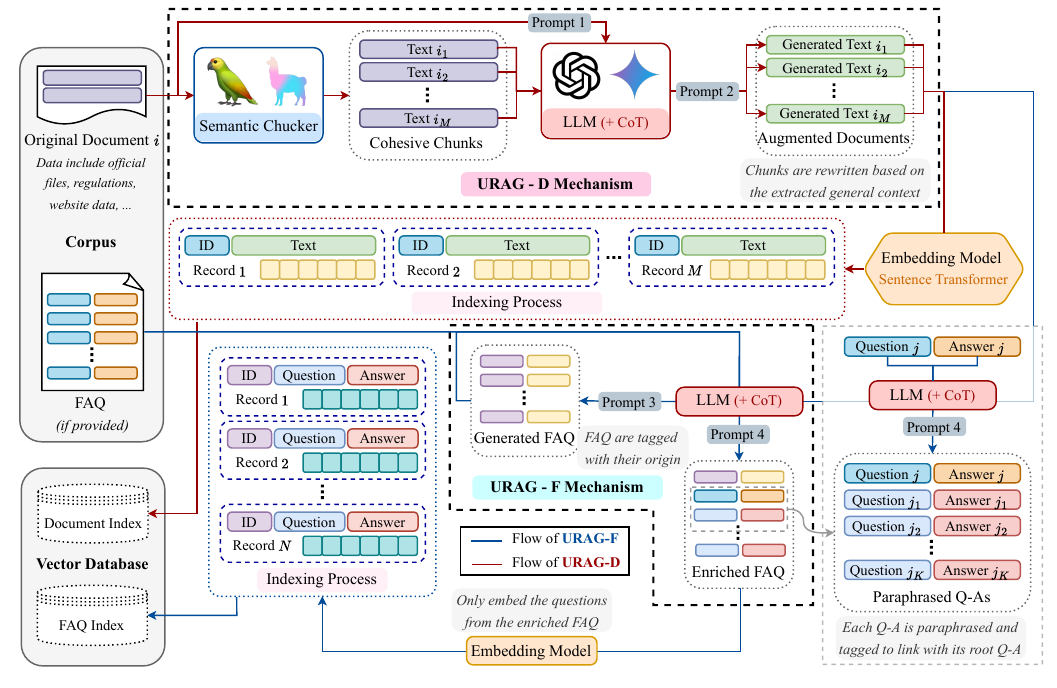}
    \caption{Illustrative overview of the two mechanisms implemented during the preparatory phase of URAG.}
    \label{URAG_Det}
\end{figure}

The URAG architecture unifies two key mechanisms during its preparatory phase to optimize performance during deployment, as shown in Figure \ref{URAG_Det}. A crucial aspect of this process is the use of \textit{Chain-of-Thought} (CoT) Prompting \cite{wei2022chain}, which enhances the reasoning capabilities of LLM-based components, enabling them to generate reliable and contextually appropriate responses. 

\medskip
\noindent \textbf{URAG-D Mechanism: Document Database Augmentation} \quad URAG-D improves document retrieval by segmenting documents into coherent chunks and strategically rewriting them for consistency and contextual relevance, instead of using entire documents as done in Naive RAG. This approach extracts the general context from the original documents, guiding the rewriting process of each chunk to maintain logical coherence. Each rewritten chunk is then condensed into a brief summary sentence, which is appended at the beginning. This workflow, detailed in Algorithm \ref{alg:URAG-D}, significantly enhances the retrieval of relevant information, thereby boosting system efficiency and accuracy.

A key component of URAG-D is \textsf{Semantic Chunking}, a novel technique proposed by \cite{semanticchunking}. Unlike traditional fixed-size chunking, semantic chunking adaptively determines chunk boundaries between sentences based on embedding similarity. This method involves analyzing the embeddings \(\mathcal{E}(s_r)\) of each sentence \( s_r \) within a document \( o_i \), and grouping sentences with similar semantic content into cohesive chunks, represented as \( c_{i_j} = \{s_r\}_r \).

Each chunk \( d_{i_j} \) is meticulously tagged to trace back to its original document \( o_i \), ensuring efficient management and retrieval within the database. In this paper, we define \( d_i = \{d_{i_j}\}_j \) as the set of processed chunks derived from the original document \( o_i \). Each \( d_{i_j} \) is treated as a distinct \textit{document} within the system, collectively forming the \textit{augmented document corpus}, denoted by \( D \).

\begin{algorithm}[!ht]
\caption{URAG-D Workflow}
\label{alg:URAG-D}
\begin{algorithmic}[1]
    \Statex \textbf{Input:} Set of original documents \( O = \{o_i\}_{i=1}^K \), where \( O \subseteq \mathbb{T} \)
    \Statex \textbf{Output:} Augmented context corpus \( D \), used in \textbf{Tier 2} of Algorithm \ref{alg:urag}
    \State Initialize \( D = \emptyset \)
    \For{\( i = 1 \) to \( K \)}
        \State \textbf{Semantic Chunking:} Apply \textsf{Semantic Chunking} on \( o_i \) to produce coherent chunks \( \{c_{i_j}\}_{j=1}^L \), where \( L=f(o_i) \) is adapted based on the content of \( o_i \)
        \State \textbf{Context Extraction:} Extract the general context \( g_i \) from \( o_i \) using \( \mathcal{L}(o_i) \) to capture overarching themes
        \Statex \textbf{Chunk Reconstruction:}
        \For{\( j = 1 \) to \( L \)}
            \State Rewrite the chunk \( c_{i_j} \) using the context \( g_i \) to form \( t_{i_j} = \mathcal{L}(c_{i_j} \mid g_i) \)
            \State Condense \( t_{i_j} \) into a succinct representation \( h_{i_j} = \mathcal{L}(t_{i_j}) \)
            \State Combine to produce the final chunk \( d_{i_j} = \text{concat}(h_{i_j}, t_{i_j}) \), where \( d_{i_j} \in \mathbb{T} \)
        \EndFor
        \State Append to the augmented corpus: \( D := D \cup \{d_{i_j}\}_j \)
    \EndFor
\end{algorithmic}
\end{algorithm}

\noindent \textbf{URAG-F Mechanism: FAQ Database Enrichment}\quad URAG-F, shown in Algorithm \ref{alg:URAG-F}, is designed to enrich the FAQ collection by fully leveraging the augmented text document corpus, obtained from the URAG-D mechanism, along with the initial FAQ set, which may contain critical Q-A pairs that require special attention. The process begins by generating new Q-A pairs from these sources and then paraphrasing them into multiple variations to enhance linguistic diversity. This approach significantly broadens and deepens the FAQ database, particularly in linguistically rich languages like Vietnamese, where a question can be articulated in numerous ways, thus improving the overall quality and coverage of the FAQ set.

\begin{algorithm}[!ht]
\caption{URAG-F Workflow}
\label{alg:URAG-F}
\begin{algorithmic}[1]
    \Statex \textbf{Input:} Context corpus \(D\) and initial FAQ set \(F_0\)
    \Statex \textbf{Output:} Enriched FAQ collection \(F\), used in \textbf{Tier 1} of Algorithm \ref{alg:urag}
    \Statex \textbf{Initial FAQ Expansion:} 
    \State Utilize \(F_0\) to generate an expanded FAQ set: \(F = F_0 \cup \mathcal{L}(F_0)\)
    \Statex \textbf{FAQ Generation from Documents:}
    \For{\(i = 1\) to \(|D|\)} 
        \State Extract Q-A pairs \(F_{D_i} = \mathcal{L}(d_i)\) from documents \(d_i\)
        \State Update the FAQ collection: \(F := F \cup F_{D_i}\)
    \EndFor
    \Statex \textbf{FAQ Enrichment:}
    \For{\(j = 1\) to \(|F|\)}
        \State Generate paraphrased variants \(\{f_{j_k}\}_k\) of each Q-A pair \(f_j\)
        \State Enrich the FAQ set with the paraphrased versions: \(F := F \cup \{f_{j_k}\}_k\)
    \EndFor
\end{algorithmic}
\end{algorithm}

\section{Experimentation} 
\label{sec:Ex}
\subsection{Experiment Setup}

We conducted an experiment to evaluate the accuracy of the URAG system in answering questions related to university admissions.

\medskip \noindent\textbf{Dataset Preparation}\quad The dataset was compiled from real questions gathered from high school students during three university admission events at HCMUT. We carefully curated 500 high-quality questions, covering a broad range of topics, emphasizing both Factual and Reasoning types. The majority of questions were Factual, reflecting common inquiries in the admissions process. To ensure a fair evaluation, we focused on questions with readily available information sourced from the official HCMUT website\footnote{\url{https://hcmut.edu.vn/}} and other reliable online platforms.

\medskip\noindent\textbf{URAG Baseline}\quad For the \textsf{Embedding Model} in our framework, we utilized the SOTA Vietnamese model \href{https://huggingface.co/dangvantuan/vietnamese-embedding}{\texttt{dangvantuan/vietnamese-embedding}}, which employs the advanced \textit{Sentence-BERT} (SBERT) \cite{Reimers2019SentenceBERTSE} architecture to enhance sentence-level semantic understanding. For \textsf{Semantic Chunking}, we employed the module provided by LangChain\footnote{\url{https://www.langchain.com/}}, a powerful framework for developing applications with LLMs. The \textsf{LLM} generator component is powered by our lightweight Vietnamese model \href{https://huggingface.co/ura-hcmut/ura-llama-7b}{\texttt{ura-hcmut/ura-llama-7b}} \cite{truong-etal-2024-crossing}, which is continually pretrained on LLaMA \cite{Touvron2023LLaMAOA} using an extensive Vietnamese dataset. In terms of data preparation, we curated 300 documents encompassing general information, events, personnel, admissions, and other relevant topics related to HCMUT. This entire setup is referred to as HCMUT Chatbot.

\medskip \noindent\textbf{Comparative Systems}\quad To benchmark URAG’s performance, we compared it against leading commercial chatbots, including GPT-4o, Claude 3.5 Sonnet, and Gemini 1.5 Pro \cite{Minaee2024LargeLM}. The evaluations were conducted through their official platforms, allowing the chatbots to fully utilize their advanced features, including Internet access and sophisticated reasoning engines. We configured them to use online search capabilities through system prompts, allowing them to retrieve additional information for questions. This mirrors the approach used in CRAG systems, where external web searches enhance response accuracy.

\subsection{Methodology}

Each question from the evaluation dataset was systematically posed to each model, and the responses were assessed by experts for correctness. Accuracy was chosen as the primary evaluation metric, calculated as shown in Equation \ref{eq:1}.

\[
\text{Accuracy} = \frac{1}{n} \sum_{i=1}^{n} \text{Correct}_i, \label{eq:1}\tag{1}
\]
where \(\text{Correct}_i\) equals $1$ if the answer to the \(i\)-th question is correct and $0$ otherwise, and \(n\) represents the total number of questions in the dataset. This metric is particularly appropriate for university admission chatbots, where the priority is on the correctness of the answers rather than their phrasing.

\setlength{\tabcolsep}{5pt} 
\renewcommand{\arraystretch}{1.3}

\subsection{Results and Analysis}

Table \ref{table:results} presents the overall accuracy of each evaluated system, showing that HCMUT Chatbot, driven by the URAG system, achieves the highest performance among all systems. This outcome highlights the effectiveness of URAG’s two-tier architecture in enhancing the accuracy of a lightweight model through efficient information retrieval management.

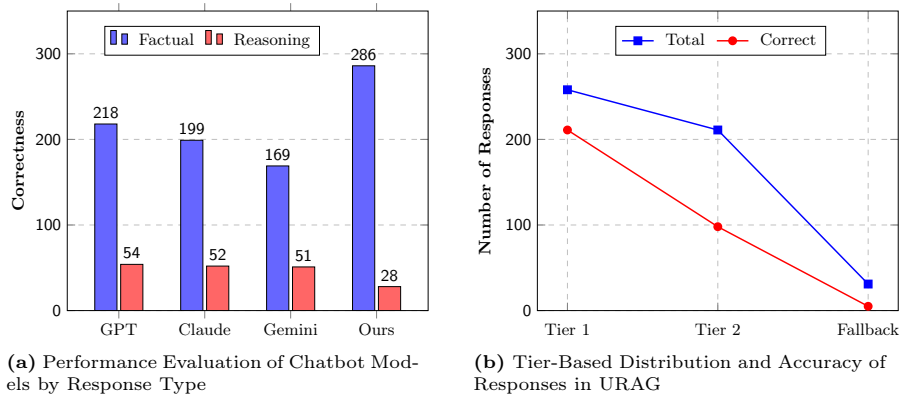
\begin{figure}[!ht]
    \centering
    \begin{subfigure}{0.45\textwidth}
        \centering
        \begin{tikzpicture}[scale=0.7] 
            \begin{axis}[
                ybar,
                bar width=12pt,
                enlarge x limits=0.2,
                legend style={at={(0.7,0.95)}, 
                    anchor=north east,legend columns=2, 
                    column sep=4pt, 
                    font=\small},
                ylabel={Correctness},
                ylabel style={font=\bfseries\sansmath},
                symbolic x coords={GPT, Claude, Gemini, Ours},
                xtick=data,
                nodes near coords,
                nodes near coords align={vertical},
                nodes near coords style={font=\sansmath}, 
                ymin=0,
                ymax=350,
                x tick label style={anchor=center, yshift=-0.2cm, font=\small\sansmath}, 
                ymajorgrids=true,
                grid style=dashed,
                tick label style={font=\small\sansmath},
            ]
            \addplot[fill=blue!60!white] coordinates {(GPT,218) (Claude,199) (Gemini,169) (Ours,286)};
            \addplot[fill=red!60!white] coordinates {(GPT,54) (Claude,52) (Gemini,51) (Ours,28)};
            \legend{Factual, Reasoning} 
            \end{axis}
        \end{tikzpicture}
        \caption{Performance Evaluation of Chatbot Models by Response Type}
        \label{fig:question_type_results}
    \end{subfigure}
    \hspace{0.5cm} 
    \begin{subfigure}{0.45\textwidth}
    \vspace{-1cm}
        \centering
        \begin{tikzpicture}[scale=0.7] 
            \begin{axis}[
                ylabel={Number of Responses},
                grid=major,
                legend style={at={(0.8,0.95)}, 
                    anchor=north east,
                    legend columns=2, 
                    column sep=4pt, 
                    font=\small},
                symbolic x coords={Tier 1, Tier 2, Fallback},
                xtick=data,
                ymin=0,
                ymax=350,
                x tick label style={anchor=center, yshift=-0.35cm, font=\small\sansmath},
                ymajorgrids=true,
                grid style=dashed,
                tick label style={font=\small\sansmath},
                ylabel style={font=\bfseries\sansmath},
                xlabel style={font=\bfseries\sansmath}, 
            ]
            \addplot[
                color=blue,
                mark=square*,
                thick
            ] coordinates {
                (Tier 1,258) (Tier 2,211) (Fallback,31)
            };
            \addplot[
                color=red,
                mark=*,
                thick
            ] coordinates {
                (Tier 1,211) (Tier 2,98) (Fallback,5)
            };
            \legend{Total, Correct}
            \end{axis}
        \end{tikzpicture}
        \caption{Tier-Based Distribution and Accuracy of Responses in URAG}
        \label{fig:tier_distribution}
    \end{subfigure}
    \caption{Comprehensive Analysis of Chatbot Model Performance and URAG's Tiers}
    \label{fig:combined_analysis}
\end{figure}

We further examined the performance based on Question Type, specifically Factual and Reasoning, as depicted in Figure \ref{fig:question_type_results}. HCMUT Chatbot performed exceptionally well in Factual questions but did not significantly surpass the other models in Reasoning tasks. This result is understandable, given that HCMUT Chatbot relies on a lightweight Vietnamese language model, whereas the commercial chatbots benefit from extensive language models with integrated reasoning engines and vast parameter counts.

To evaluate the impact of each phase of the URAG architecture, we analyzed the distribution and accuracy of responses across its tiers, as shown in Figure \ref{fig:tier_distribution}. The findings indicate that the URAG-F and URAG-D mechanisms were highly effective, addressing nearly all questions within the first two tiers while keeping fallback cases minimal. There was a balanced distribution of responses between Tier 1 and Tier 2, with a slight decline in accuracy as the system progressed through the tiers, reflecting the designed hierarchy of the retrieval process.

\begin{table}[!ht]
    \centering
    \caption{Performance Metrics of Evaluated Chatbot Systems}
    \label{table:results}
    \begin{tabular}{lcc}
        \toprule
        \textbf{System} & \textbf{Correctness} & \textbf{Accuracy} \\
        \midrule
        GPT-4o & 272 & 0.544 \\
        Claude 3.5 Sonnet & 251 & 0.502 \\
        Gemini 1.5 Pro & 220 & 0.440 \\
        HCMUT Chatbot & \textbf{314} & \textbf{0.628} \\
        \bottomrule
    \end{tabular}
\end{table}
\vspace{-0.5cm}

\subsection{Case Study: Deployment of the HCMUT Admission Chatbot}

\begin{figure}[!ht]
    \centering
    \includegraphics[width=\linewidth]{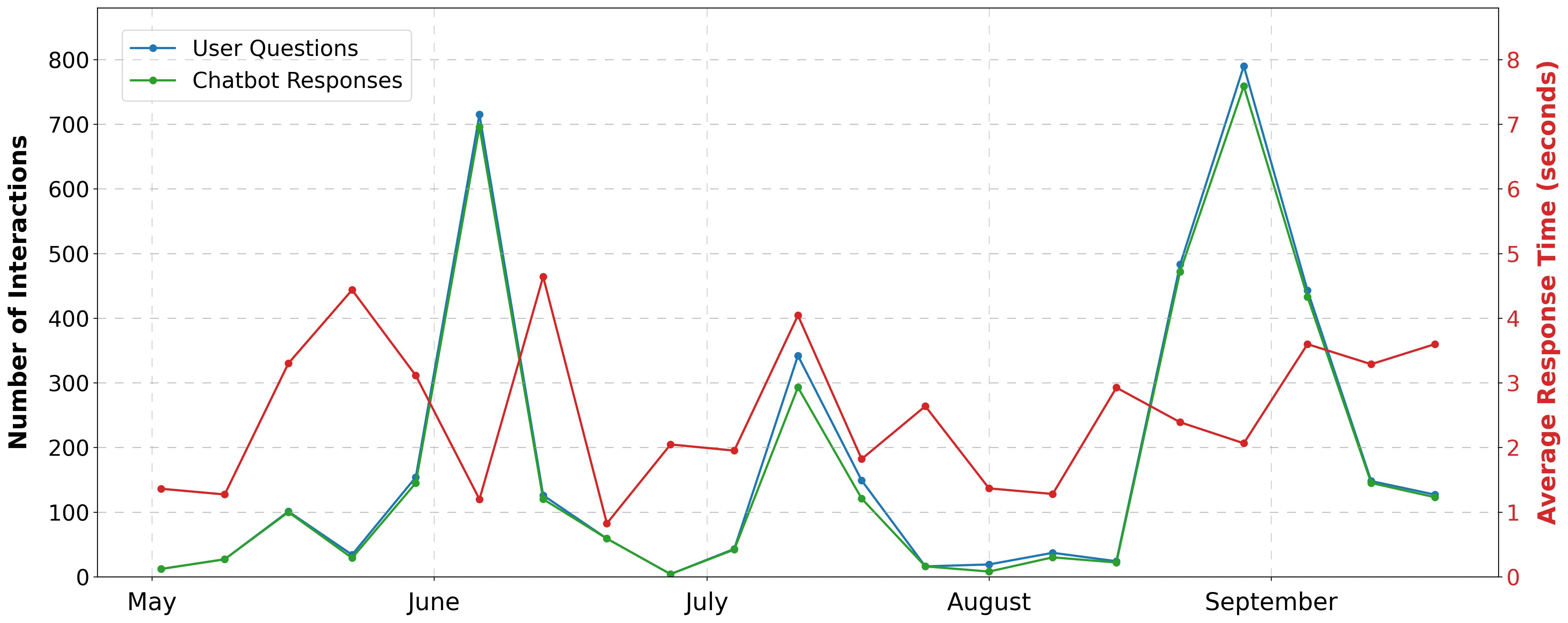}
    \caption{Interaction and Response Time Statistics of HCMUT Chatbot}
    \label{fig:practical}
\end{figure}

We deployed HCMUT Chatbot, powered by the entire URAG baseline, on a dedicated subdomain of our university’s website at \href{https://www.ura.hcmut.edu.vn/bk-tvts/}{\texttt{ura.hcmut.edu.vn/bk-tvts}}. Over a four-month deployment period, the chatbot recorded substantial interaction levels, particularly from high school students, as shown in Figure \ref{fig:practical}. Interaction peaks were observed in early June, coinciding with the end of the high school academic year, and late August, just before the start of the university term, aligning with the expected demand for admission information. Our chatbot maintained fast response times that were consistent across varying interaction volumes, with slight variations depending on the complexity of the questions.

\section{Discussion, Limitations, and Future Work}

We introduced the URAG system as an efficient and lightweight solution for university admission chatbots, successfully deployed at HCMUT with a robust and stable user base. Its two-tier architecture significantly enhances accuracy, particularly for high-priority questions, by mitigating common hallucination issues inherent in LLM-based systems. The traceability features in URAG-D and URAG-F efficiently manage a large enriched database by linking text and question variations to their original sources, simplifying updates when information changes. However, URAG’s reliance on a lightweight generator model, while advantageous for reducing deployment costs and enhancing security, limits its performance compared to larger models or third-party APIs, especially when addressing general queries beyond the admissions domain \cite{truong-etal-2024-crossing}.

Future research should focus on advanced retrieval strategies, such as integrating SOTA Hybrid Search techniques \cite{Wang2024SearchingFB}, to replace the current Cosine Similarity with Threshold method, thereby improving retrieval efficiency. Additionally, fine-tuning the LLM on domain-specific datasets related to university admissions could further enhance the relevance and accuracy of responses. However, these improvements must be carefully balanced against the potential increase in operational complexity and costs. Lastly, it is essential to acknowledge that certain inquiries will always require the expertise and judgment of human advisors, which automated systems cannot fully replicate.

\bibliographystyle{ieeetr}
\bibliography{references}


\appendix \section{Appendix}
\subsection{Evaluation Dataset Analysis}
\setlength{\tabcolsep}{3pt} 
\renewcommand{\arraystretch}{1.3}
The evaluation dataset comprises 500 authentic questions collected from high school students, covering a wide range of topics and including both Factual and Reasoning question types. The detailed composition of the dataset is presented in Table \ref{table:combined}. 

\begin{table}[!ht]
    \centering
    \caption{Distribution of Questions by Topic and Question Type}
    \label{table:combined}
    \resizebox{\textwidth}{!}{%
    \begin{tabular}{@{}llp{8.8cm}r@{}}
        \toprule
        \textbf{Category} & \textbf{Type} & \textbf{Description} & \textbf{Count} \\
        \midrule
        \multirow{5}{*}{\raisebox{-11ex}[0pt][0pt]{Topic}} 
        & University Overview & General inquiries about the university’s founding, locations, facilities, organizational structure, and overall campus environment. & 61 \\
        & Programs and Majors & Related to academic programs, entry requirements, curriculum details, unique aspects of majors, and opportunities for internships and research. & 186 \\
        & Faculty and Career Guidance & Inquiries about faculty expertise, teaching and research experience, as well as guidance on career opportunities and job placement for students. & 32 \\
        & Student Life and Extracurriculars & About student clubs, extracurricular activities, events, sports, student welfare, and support services. & 50 \\
        & Admissions and Policies & On admission criteria, application procedures, scholarships, financial aid, health insurance, and other student policies. & 171 \\
        \midrule 
        \multicolumn{3}{r}{\textbf{Total}} & \textbf{500} \\
        \midrule
        \multirow{2}{*}{\raisebox{-4.5ex}[0pt][0pt]{Question Type}} 
        & Factual & Seeking specific information or details about programs, procedures, admission requirements, facilities, and services offered by the university, based on factual data. & 427 \\
        & Reasoning & Require analysis, comparison, explanation, or advice regarding academic choices, program distinctions, and career guidance. & 73 \\
        \bottomrule
    \end{tabular}%
    }
\end{table}

\subsection{Hyperparameter Selection for HCMUT Chatbot}
\setlength{\tabcolsep}{5pt} 
\renewcommand{\arraystretch}{1.3}
As described in Algorithm \ref{alg:urag}, selecting the appropriate thresholds \( t_{\text{FAQ}} \) and \( t_{\text{Doc}} \) for Tier 1 and Tier 2, respectively, is crucial to optimizing the performance of the URAG system. To fine-tune these hyperparameters, we conducted experiments using 500 randomly generated queries based on data from the FAQ set \( F \) and document set \( D \), processed through the URAG-D and URAG-F mechanisms.

\medskip
\noindent\textbf{Threshold Selection for \( t_{\text{FAQ}} \)}\quad  
To determine the optimal value of \( t_{\text{FAQ}} \), we retrieved the top \( k \) most relevant questions from the FAQ set \( F \) for each query. The performance was evaluated using the \textit{Mean Reciprocal Rank} (MRR), computed as shown in Equation \ref{eq:2}.
\[
\text{MRR} = \frac{1}{|Q|} \sum_{i=1}^{|Q|} \frac{1}{\text{rank}_i}, \label{eq:2} \tag{A.2}
\]
where \(\text{rank}_i\) is the position of the first relevant question retrieved for the \(i\)-th query, and \(\text{rank}_i = \infty\) if no relevant question is found. The total number of queries is \(|Q| = 500\). MRR was selected because it effectively captures the quality of retrieval by emphasizing the rank of the first correct result, providing a robust measure of retrieval system performance, particularly suited to the nature of URAG's tiers. By testing threshold values between \(0.8\) and \(0.95\), we identified the \( t_{\text{FAQ}} \) that maximized MRR. The value \( k = 20 \) was chosen to align with the average number of paraphrased variations generated from a single original object by the URAG-F mechanism, ensuring consistency and meaningful evaluation.

\medskip
\noindent\textbf{Threshold Selection for \( t_{\text{Doc}} \)}\quad  
The approach for selecting \( t_{\text{Doc}} \) was similar, focusing on retrieving the most relevant document segments from the document set \( D \). We set \( K = 2 \), striking a balance between computational efficiency and retrieval accuracy, tailored to the capabilities of our lightweight LLM. 

\medskip
The final hyperparameters used in Section \ref{sec:Ex} are summarized in Table \ref{table:hyperparams}.

\begin{table}[!ht]
    \centering
    \caption{Hyperparameters for URAG Components}
    \label{table:hyperparams}
    \begin{tabular}{@{}llc@{}}
        \toprule
        \textbf{Stage} & \textbf{Hyperparameter}         & \textbf{Value} \\ \midrule
        \multirow{2}{*}{\raisebox{0.2ex}[0pt][0pt]{Tier 1}} & \verb|THRESHOLD_FAQ| & 0.9  \\ 
                                                             & \verb|TOP_K|         & 20   \\ \midrule
        \multirow{2}{*}{\raisebox{0.2ex}[0pt][0pt]{Tier 2}} & \verb|THRESHOLD_DOC| & 0.8  \\ 
                                                             & \verb|TOP_K|         & 2    \\ \midrule
        \multirow{4}{*}{\raisebox{0.2ex}[0pt][0pt]{URA-LLaMA}} & \verb|temperature| & 0.9  \\ 
                                                             & \verb|top_p|         & 0.95 \\ 
                                                             & \verb|top_k|         & 40   \\ 
                                                             & \verb|max_new_tokens|& 512  \\ \bottomrule
    \end{tabular}
\end{table}


\end{document}